\pdfoutput=1

\documentclass[11pt]{article}

\usepackage[final]{acl}

\usepackage{times}
\usepackage{latexsym}
\usepackage{amsthm}
\usepackage{amsmath,amssymb}
\usepackage[ruled,linesnumbered]{algorithm2e}
\DeclareMathOperator*{\argmax}{arg\,max} 
\usepackage{booktabs}
\usepackage[T1]{fontenc}

\usepackage[utf8]{inputenc}

\usepackage{microtype}

\usepackage{inconsolata}

\usepackage{graphicx}

\title{On the Fragility of Active Learners for Text Classification}

\author{Abhishek Ghose \\
    {[24]}7.ai \\
  \texttt{abhishek.ghose.82@gmail.com} \\\And
  Emma Thuong Nguyen \\
  {[24]}7.ai \\
  \texttt{emma.ttn8@gmail.com } \\}

\begin{document}
\maketitle
\begin{abstract}

Active learning (AL) techniques optimally utilize a labeling budget by iteratively selecting instances that are most valuable for learning. However, they lack ``prerequisite checks'', i.e., there are no prescribed criteria to pick an AL algorithm best suited for a dataset. A practitioner  must pick a technique they \emph{trust} would beat random sampling, based on prior reported results, and hope that it is resilient to the many variables in their environment: dataset, labeling budget and prediction pipelines. The important questions then are: how often on average, do we expect any AL technique to reliably beat the computationally cheap and easy-to-implement strategy of random sampling? Does it at least make sense to use AL in an ``Always ON'' mode in a prediction pipeline, so that while it might not always help, it never under-performs random sampling?  How much of a role does the prediction pipeline play in  AL's success?

We examine these questions in detail for the task of text classification using pre-trained representations, which are ubiquitous today.

Our primary contribution here is a rigorous evaluation of AL techniques, old and new, across setups that vary wrt datasets, text representations and classifiers. This unlocks multiple insights around warm-up times, i.e., number of labels before gains from AL are seen, viability of an ``Always ON'' mode and the relative significance of different factors.
Additionally, we release a  framework for rigorous benchmarking of AL techniques for text classification.


 
 
\end{abstract}

\section{Introduction}

Within a supervised learning setup, Active Learning (AL) techniques \cite{settles2009active} use a \emph{Query Strategy (QS)} to identify an unlabeled set of instances which is optimal in the following sense: if labelled and added to the training data, they lead to the greatest improvement in model accuracy, relative to any other same-sized set. In cases where labelling is expensive, the value proposition of AL is that it is cost-efficient compared to \emph{random sampling}, and a model reaches greater accuracy with a smaller number of labelled instances.  

In practice, an AL technique is selected based on the strength of prior reported results, i.e., there are no ``prerequisite checks'':  tests that one might perform on an unlabeled dataset, that help to select a  technique suited to a problem\footnote{We refer to this as the practitioner's \emph{decision model} and formalize it in \S \ref{sec:decision_model}.}. This trust extends to related decisions such as batch and seed sizes, as well as the hyperparameters (if any) of the AL technique since there is no way to empirically pick them: to compare with random sampling, or among techniques, labels are required. But if one had labels, they wouldn't need AL \cite{10.1145/1964897.1964906}! In this sense, the AL setup is unforgiving as one needs to make the optimal choice in one shot \cite{margatina-aletras-2023-limitations}.

This leads us to ask multiple questions about the broader area. How valid is  the implicit but consequential assumption of \emph{transferability}? A related question is whether the focus on QSes alone is warranted - how much do the other components of a prediction pipeline affect outcomes? And finally, does it make sense to use AL at least in an ``Always ON'' mode in a data labeling workflow; this is akin to asking if AL might perform \emph{worse} than random sampling. We need to quantify both the frequency and magnitude of gains from AL, to be able to evaluate the cost of such pipelines. This is because even simple AL techniques require a model to be evaluated over the unlabeled data pool, which can be expensive depending on the model complexity, size of the data pool and the latency allowed per AL iteration.

To be clear, we don't question if AL results are reproducible within the \emph{original setups} they were reported in\footnote{In the interest of fairness, we conducted limited \textbf{reproduciblity tests} for the AL techniques we benchmark here, and were able to replicate reported results - see \S \ref{sec:app_repr}. }; but whether any of those gains carry forward to \emph{new setups}, which is how AL is used in practice.

We pick the area of text classification to investigate these concerns. The larger area of NLP has seen a rapid infusion of novel ideas of late. Today, a practitioner has easy access to a variety of powerful classifiers via packages such as \emph{scikit-learn} \cite{scikit-learn}, \emph{spaCy} \cite{spacy} and \emph{Hugging Face} \cite{wolf-etal-2020-transformers}, and text representations, such as \emph{Universal Sentence Encoding (USE)} \cite{USE}, \emph{MiniLM} \cite{minilmv2} and \emph{MPNet} \cite{mpnet}. This makes it a fertile ground for testing AL's utility.

In all this, \emph{our motivation is not to disapprove of AL as an area for research, but to motivate the inclusion of multiple practical challenges in future studies}.

\textbf{Contributions}: Our primary contribution is a rigorous empirical analysis of the learning behavior of AL techniques over multiple text classification pipelines, that is targeted towards answering the questions asked above. Additionally,  we open source an AL evaluation framework\footnote{Our framework, \emph{ALchemist}, is available here: \url{https://github.com/ThuongTNguyen/ALchemist}.}, to enable researchers to not only reproduce our analysis, but also to rigorously evaluate their own contributions.

\section{Previous Work}
\label{sec:prev_work}
Critique of AL is not new. \citet{10.1145/1964897.1964906} criticize AL for its unpredictable (for a task) warm-up times, i.e., a minimum number of labeled instances before which gains over random sampling are evident. \citet{margatina-aletras-2023-limitations} point out problems with AL simulations. \citet{navigating} identify key issues leading to a lack of realistic AL evaluations and propose solutions that they apply to image classification. \citet{lowell-etal-2019-practical} study AL empirically but focus on the interesting notion of \emph{successor models}, i.e., future models that would use the labeled data collected via AL using a specific model. \citet{ijcai2021p634} examine the empirical effectiveness of AL, but they don't evaluate on NLP tasks. \citet{siddhant-lipton-2018-deep} is an empirical study of AL effectiveness similar in spirit to ours, but they focus on deep Bayesian methods. \citet{prabhu-etal-2019-sampling} study sampling biases in deep AL, but their study is limited to one prediction model - \emph{FastText.zip} \cite{DBLP:journals/corr/JoulinGBDJM16} - and considers only QSes based on \emph{uncertainty sampling}.

This work differs from from existing literature wrt being a combination of: focusing on text classification, being empirical, employing a breadth of models (traditional and deep learning based) and employing recent techniques, e.g., \emph{MPNet} \cite{mpnet}, \emph{REAL} \cite{REAL}. While some conclusions we draw here might be similar to those reported earlier, we note that it is important to revise our collective mental models in a fast evolving area such as NLP, and in enabling that, even such conclusions are valuable.



\section{Batch Active Learning - Overview}
\label{sec:bal_overview}
In this work, we specifically study the \emph{batch} AL setting for text classification. Here, a QS identifies a \emph{batch} of $b$ unlabeled points, at each iteration $t$, for $T$ iterations. A model $M_t$, that is trained on the accumulated labeled pool, is produced at the end of each iteration. The first iteration uses a seed set of $s$ randomly sampled points (although other strategies may be used).

We note that that $M_t$ should be produced using a \emph{model selection} strategy (we use a hold-out set here), and must also be \emph{calibrated} (we use \emph{Platt scaling} \cite{platt_calib,calib_compare}). The former ensures that $M_t$ doesn't overfit to the labeled data, which is likely in the initial iterations due to small quantities. The latter is required since many query strategies rely on uncertainty/confidence scores produced by $M_t$. Unfortunately, in our experience, multiple implementations/studies miss one or both of these steps.

To avoid any ambiguity, we provide pseudo-code for this AL setting in Algorithm \ref{algo:sketch} in \S \ref{sec:appendix_pseudocode_bal}.

\section{Experiment Setup}
\label{sec:experiment_setup}
In this section, we describe our experiment setup in detail.

\subsection{Configuration Space of Experiments}
\label{sec:method}
\begin{figure*}[ht]
\includegraphics[width=\textwidth]{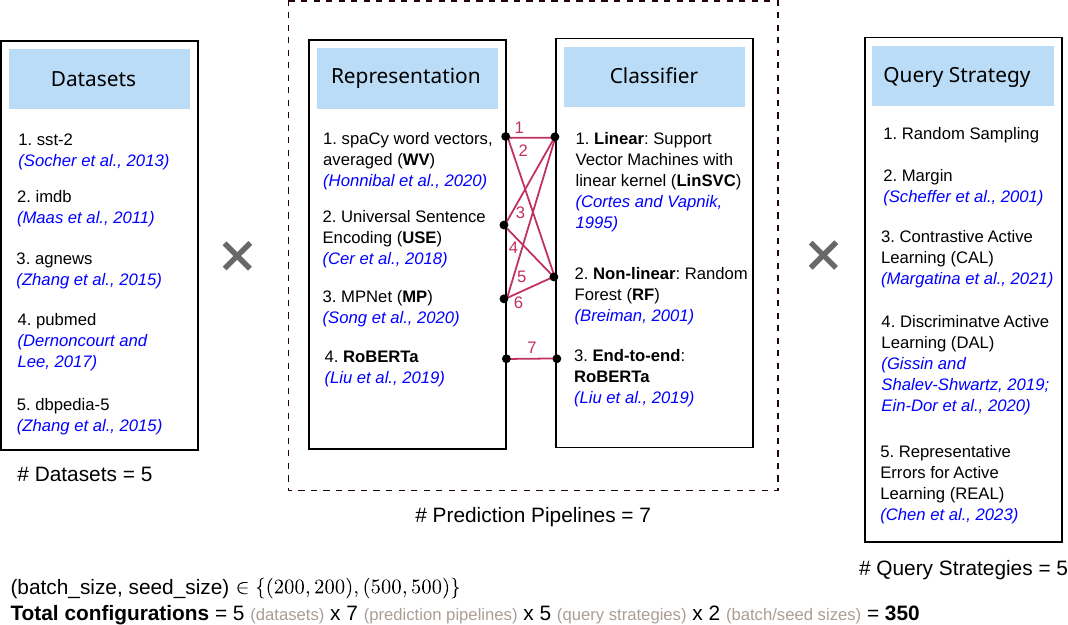}
\caption{The space of experiments is shown. See \S \ref{sec:method} for description. All representations are produced by pre-trained models, which are ubiquitous in practice today. The lines between the boxes ``Representation'' and ``Classifier'' denote combinations that constitute our prediction pipelines. Note that RoBERTa is an end-to-end predictor, where there are no separate representation and classification steps. Also note that the popular \emph{Transformer} architecture \cite{NIPS2017_3f5ee243} is represented by RoBERTa and MPNet here.} \label{fig:experiment_space}
\end{figure*}
Our experiment configurations vary wrt \emph{datasets}, \emph{text representations}, \emph{classifiers}, the \emph{batch} and \emph{seed sizes}, and of course, the \emph{QS}. We study the following QS here: (1) \emph{Random} as baseline, (2) \emph{Margin}\footnote{Also referred to as \emph{Smallest Margin} or \emph{Breaking Ties}, it is still considered to be competitive \cite{schroeder2022-revisiting}.} \cite{10.1007/3-540-44816-0_31,schroeder2022-revisiting}, (3) \emph{Contrastive Active Learning (CAL)} \cite{cal}, (4) \emph{Discriminative Active Learning (DAL)} \cite{gissin2019discriminative,ein-dor-etal-2020-active}, and (5) \emph{Representative Errors for Active Learning (REAL)} \cite{REAL}.
We picked these  either because they are contemporary, e.g., \emph{REAL}, \emph{DAL}, \emph{CAL}, or have produced strong contemporary results, e.g., \emph{Margin}.

Figure \ref{fig:experiment_space} enumerates the configuration space. For further details (including hyperparameters) see \S \ref{sec:appendix_experiment_space} and \S \ref{sec:app_hyp}. Note that all representations used are based on \emph{pre-trained} models which have grown quite popular in the past few years. For classification, we picked one each of a linear, non-linear and Deep Learning based classifier\footnote{Although end-to-end classifiers, e.g., RoBERTa, DistilBERT \cite{sanh2020distilbertdistilledversionbert}, are popular today, we include pipelines with separate representation and classification components since they are still used where: (a) a good latency-accuracy trade-off is needed, and (b) there are multiple downstream tasks that might leverage the representation, e.g., classification, similarity-based retrieval, sentiment analysis. On a different note, the growing popularity of \emph{Retrieval Augmented Generation (RAG)} \cite{NEURIPS2020_6b493230} has re-shifted focus to the area of learning good embeddings.}. Since batch or seed sizes are inconsistent in AL literature, e.g., DAL, REAL and CAL respectively use batch sizes of $50$, $150$, $2280$ - we vary these settings as well.

\begin{figure*}[t]
\includegraphics[width=\textwidth]{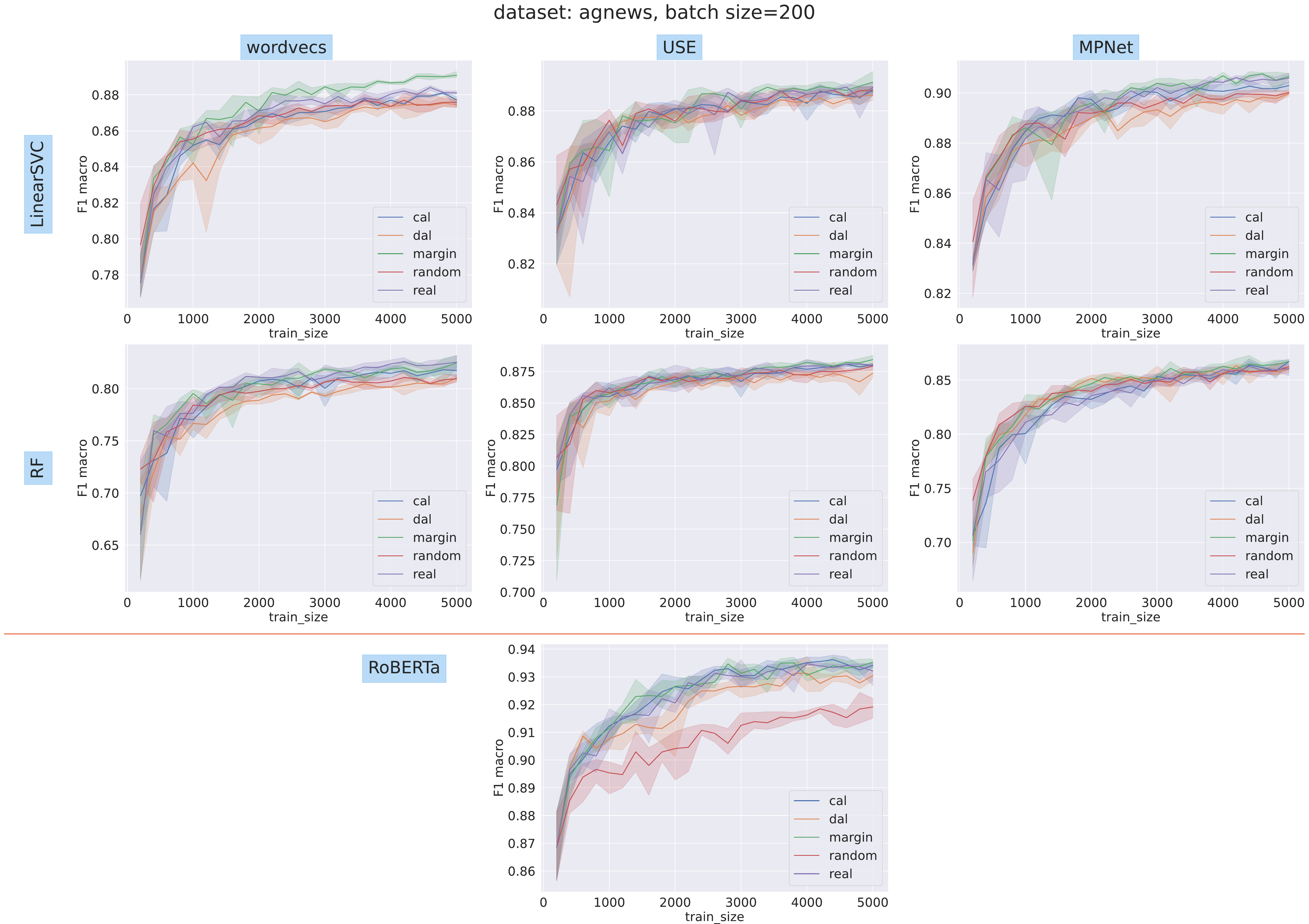}
\caption{F1 macro scores on the test set at each iteration, for the dataset \emph{agnews} and batch size of $200$. The $x$-axes show size of the labeled data, the $y$-axes show the F1-macro scores on the test data. } \label{fig:datawise_sample}
\end{figure*}

For an idea of the breadth of this search space, see Figure \ref{fig:datawise_sample} which shows results for the dataset \emph{agnews} and batch/seed size of $(200, 200)$.

\subsection{Metrics and Other Settings}
\label{sec:metrics_other_settings}
The \textbf{classifier accuracy metric} we use is the \textbf{F1 (macro)} score, since it prevents performance wrt dominant classes from overwhelming results. For measuring the \textbf{effectiveness of a QS},we use the \textbf{relative improvement wrt the random QS} of the classifier score (see Equation \ref{eqn:rel_improve}). The \textbf{size of the unlabeled pool} is $\boldsymbol{20000}$ at the start of each experiment. If the original dataset has more than than $20000$ instances, we extract a label-stratified sample, to retain the original class distribution. The \textbf{size of the test set} is $\boldsymbol{5000}$ - also a label-stratified sample from the corresponding test set of the original dataset. 

We run an experiment till the size of the labeled set has grown to $5000$ instances\footnote{Beyond this labeled set size (unrelated to the test set size) different QSes produce similar gains - see \S \ref{sec:app_variance}.}. This implies $\boldsymbol{T}=(5000 - 200)/200 \boldsymbol{= 24}$ \textbf{iterations} for the batch/seed size setting of $(200, 200)$, and similarly $\boldsymbol{T=9}$ \textbf{iterations} for the $(500, 500)$ setting.

\;
As shown in Figure \ref{fig:experiment_space} we have $\boldsymbol{350}$ \textbf{unique configurations}. We also execute \textbf{each configuration three times} in the interest of robust reporting. This gives us a a total of  $350 \times 3=\boldsymbol{1050}$ \textbf{trials}. For \emph{each AL iteration} of \emph{each of these trials}, we follow the due process of model selection\footnote{\citet{margatina-aletras-2023-limitations} point out that this is lacking in most AL studies. This is another way the current work differentiates itself.} and calibration\footnote{\emph{RoBERTa} is the only exception since it is naturally well-calibrated \cite{desai-durrett-2020-calibration}.}.  



\subsection{Notation and Terminology}
\label{sec:notation}

We introduce some notation here that will help us precisely describe our analysis in later sections.

Let $f$ be a function that computes the model metric of interest, e.g., \emph{F1-macro}. This  accepts, as parameters, the \emph{random variables}\footnote{Of course, in this work we consider them to only assume  specifically chosen values, e.g., RF, LinSVC and RoBERTa as  predictors.} $h, q, d, b, s, n$, which are defined as follows:

\begin{itemize}
    \item $h \in H$, the set of prediction pipelines.
    \item $q \in Q$, the set of query strategies.  For convenience, we also define $q_R$ to be the \emph{random} QS, and $\mathcal{Q}_{NR} = \{cal, dal, real, margin\}$, i.e., the subset of non-random QS. 
    \item $d\in D$, the set of datasets.
    \item $(b,s) \in V$, the set of batch and seed size combinations, i.e., $V=\{(200, 200), (500, 500)\}$
    \item $n$ is the size of the labeled data. In our experiments,  $s \leq n \leq 5000$.
\end{itemize}
A specific value is indicated with a prime symbol on the corresponding variable, e.g., $h'$ is a specific prediction pipeline.

\textbf{QS Effectiveness}: We evaluate a non-random QS by measuring the relative improvement wrt the random QS, at a given number of labeled instances $n'$. We use the shorthand $\delta$: 
\begin{align}
    &\delta(f(h, q, d, b, s, n')) = 100 \times \nonumber \\ 
    &\frac{f(h, \boldsymbol{q}, d, b, s, n')- f(h, \boldsymbol{q_R}, d, b, s, n')}{f(h, \boldsymbol{q_R}, d, b, s, n')}
\label{eqn:rel_improve}
\end{align}

\subsection{Decision Model}
\label{sec:decision_model}
Before looking at the results, we formalize the \emph{decision model} of a practitioner using our notation. This helps us justify the  aggregations we perform over results of individual experiments.

Because of lacking prerequisite checks, there is no preference for picking  a factor in combination with others. We model them as \emph{independent} variables, i.e., the probability of a configuration is $p(h)p(q)p(d)p(b, s)$. Since each of these probabilities is also \emph{uniform}, e.g., the general practitioner is equally likely to encounter any dataset $d \in D$, each configuration has an identical probability of occurrence\footnote{They may inherit an environment with a specific prediction pipeline or a query strategy - we also present these conditional results. But within these conditions, the other factors are assumed to be independent and individually uniform.}: $1/(|H|\times|Q|\times|D|\times|V|)$. In other words, any \emph{expectation} we wish to compute over these settings under this decision model is a simple \emph{average}.

\section{Results}
\label{sec:results}
We are now ready to look at the results of our experiments.

\subsection{Expected Gains from AL}
\label{sec:expected_gains}

\begin{figure*}[t]
\begin{center}
\includegraphics[width=1\textwidth]{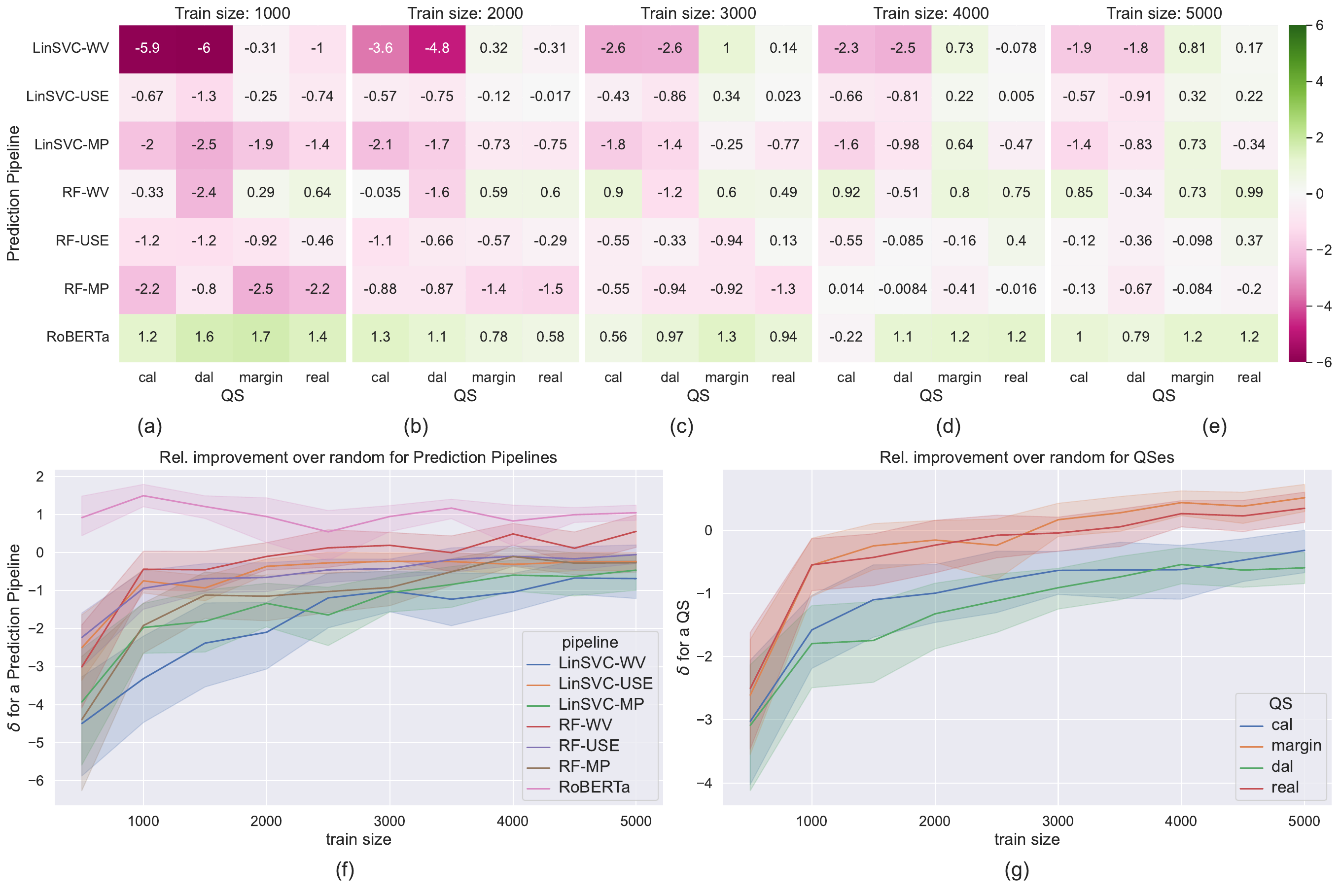}
\caption{Expected relative improvement in \emph{F1-macro} score over random. (a)-(e) show this for different predictors and QS, at different training sizes (see titles). These correspond to Equation \ref{eqn:avg_gain}. (f) and (g) show marginalized improvements for different predictors and QSes respectively; see equations \ref{eqn:avg_gain_pipelines} and \ref{eqn:avg_gain_QS}. } \label{fig:combined_rel_improve}
\end{center}
\end{figure*}
Figure \ref{fig:combined_rel_improve} shows the expected relative improvement, grouped in the following ways:
\begin{enumerate}
    \item Figure \ref{fig:combined_rel_improve}(a)-(e): These heatmaps show the expected $\delta$ at a given number of instances $n' \in \{1000, 2000, 3000, 4000, 5000\}$. A cell for predictor $h'$ and a QS $q' \in Q_{NR}$ in the heatmap for $n'$ training instances shows\footnote{This expectation is over batch and seed sizes at given values of $n'$; but note, different batch sizes \emph{don't produce same values for $n'$}. This is explicitly reconciled - see \S \ref{sec:app_batch_size_avg}.}:
        \begin{equation}
            \mathbb{E}_{d, b,s }[\delta(f(h', q', d, b, s, n'))]
        \label{eqn:avg_gain}
        \end{equation}
    The rows are arranged roughly in increasing order of classifier capacity, i.e., \emph{LinSVC}, \emph{RF}, \emph{RoBERTa}, and within a group, in increasing order of approximate representation quality: word vectors (\emph{WV}), \emph{USE}, \emph{MPNet}\footnote{The relative ordering of USE vs MPNet was obtained from the \emph{Massive Text Embedding Benchmark (MTEB)} \href{https://huggingface.co/spaces/mteb/leaderboard}{rankings}, where \emph{MPNET} leads \emph{USE} by $\sim 100$ positions today.}.   
    \item  Figure \ref{fig:combined_rel_improve}(f): This shows $\delta$ only for prediction pipelines, marginalizing over QSes. This is easy to show in a standard line-plot. The $y$-value for $x=n'$ for predictor $h'$ denotes:
        \begin{equation}
            \mathbb{E}_{d, b,s, q \in Q_{NR} }[\delta(f(h', q, d, b, s, n'))]
        \label{eqn:avg_gain_pipelines}
        \end{equation}
    
    \item Figure \ref{fig:combined_rel_improve}(g): This is analogous to (f) and shows $\delta$ for QSes while marginalizing over predictors. The $y$-value for a specific $x=n'$ for QS $q' \in Q_{NR}$ denotes: 
        \begin{equation}
            \mathbb{E}_{d, b,s, h }[\delta(f(h, q', d, b, s, n'))]
        \label{eqn:avg_gain_QS}
        \end{equation}
\end{enumerate}

\emph{Observations}: In Figure \ref{fig:combined_rel_improve}(a)-(e), we see that as we move towards the right, the number of cells with $\delta \gtrapprox 0$  increases. This suggests that, in general, as the pool of labeled instances grows, AL becomes more effective. This might seem promising at first, but note that (a) we cannot predict \emph{when} this happens in practice: we lack the theoretical tools, and it varies wrt both the predictor and the QS, and (b) if you look closely, its not that AL is becoming more effective but, rather, all configurations are converging towards\footnote{This is something we observe in a separate analysis as well - see \S \ref{sec:app_variance}. In fact, this is the reason why we grow the labeled set to only $5000$ instances in our experiments - mentioned in \S \ref{sec:app_variance}.} $\delta=0$. In other words, \emph{in low label regimes, where we expect AL to benefit us, there can be a lot of variance - it might even under-perform random sampling - and at high label regimes, their performance, even if positive, is not very different from random sampling}.

Among predictors (Figure \ref{fig:combined_rel_improve}(f), but this is also apparent in (a)-(e)), for \emph{RoBERTa} we consistently observe $\delta>0$. But we note that this value isn't high, i.e., $\delta \approx 1$. Among QSes, \emph{REAL} and \emph{Margin}, seem to do well at larger data regimes - as visible in Figure \ref{fig:combined_rel_improve}(g), but also in (d) and (e). The performance of \emph{Margin} might seem somewhat surprising, since this is an old technique (proposed in  \citet{10.1007/3-540-44816-0_31}), but similar observations have been  reported elsewhere \cite{schroeder2022-revisiting}.

\subsection{Always ON Mode}
\label{sec:always_on}
Another question we might ask is that even if AL doesn't always surpass random, is there a downside to making it a permanent part of a labeling workflow - multiple tools allow this today\footnote{\textbf{Important}: We have \emph{not} evaluated these tools. They are cited as examples of common tools used in data labeling workflows in the industry.}, e.g., \citet{prodigy_montani_honnibal,LabelStudio}? 

Table \ref{tab:always_on} shows some relevant numbers.

\begin{table}[!ht]
    \centering
    \begin{tabular}{lrrr}
    \toprule
        \emph{Avg. for} & \% times $\delta < 0 $ & $\overline{\delta}_{\geq 0}$ & $\overline{\delta}$ \\ \toprule
        \textbf{Overall} & 51.82 & 0.89 & -0.74 \\ \midrule
        LinSVC-WV & 61.71 & 0.70 & -1.90 \\ 
        LinSVC-USE & 61.57 & 0.46 & -0.64 \\ 
        LinSVC-MP & 63.71 & 0.40 & -1.48 \\ 
        
        RF-WV & 47.29 & 1.31 & -0.30 \\ 
        RF-USE & 60.57 & 0.71 & -0.63 \\ 
        RF-MP & 60.14 & 0.60 & -1.24 \\ 
        
        RoBERTa & 7.71 & 1.29 & 1.01 \\ 
        \midrule
        CAL & 55.60 & 0.81 & -1.07 \\ 
        DAL & 70.12 & 0.82 & -1.29 \\ 
        Margin & 38.45 & 0.97 & -0.25 \\ 
        REAL & 43.10 & 0.89 & -0.34 \\ \bottomrule
    \end{tabular}
      \caption{The $\%$-age of times model \emph{F1-macro} scores are worse than random are shown. Also shown are the average $\delta$s when scores are at least as good as random, and average $\delta$s in general. These are relevant to the ``Always ON'' mode, discussed in \S \ref{sec:always_on}. See Table \ref{tab:always_on_std} in \S \ref{sec:app_always_on} for standard deviations.}
\label{tab:always_on}
\end{table}

 \emph{Observations}: In general, (first row, ``\textbf{Overall}''),  the number of incidents where the relative improvement was \emph{strictly negative} (counted at various labeled data sizes across configurations) is $51.82\%$. This might be suggested by the heatmaps in Figure \ref{fig:combined_rel_improve}(a)-(e) as well, where approximately the left upper triangle of the plots combined indicates $\delta < 0$. The average improvement when AL is as good as random is low, i.e., $\overline{\delta}_{\geq 0} = 0.89$, and on the whole this quantity is \emph{actually negative}, i.e., $\overline{\delta}=-0.74$. Again, the use of \emph{RoBERTa} leads to favorable scores. Among QSes, \emph{Margin} and \emph{REAL} perform relatively well.
 
 Under our decision model - \S \ref{sec:decision_model} - the practical implication is bleak: in the ``Always ON'' mode, stopping labeling early risks negative improvement. The only way to ensure $\delta \geq 0$ is to accumulate quite a few labels, i.e., move out of the left upper triangular region in Figure \ref{fig:combined_rel_improve}(a)-(e), but then the average improvement is low. Essentially, the ``Always ON'' mode is viable if the small relative gains from labeling $4000{-}5000$ instances are useful. 

\subsection{Effect of Prediction Pipeline vs QS}
\label{sec:pipeline_or_QS}
Papers on AL typically contribute QSes. Here we ask if that focus is warranted, i.e.,  what has a greater impact? - the QS or the prediction pipeline?

We might suspect that it is  the pipeline, given the performance of \emph{RoBERTa} in both Figure \ref{fig:combined_rel_improve} and Table \ref{tab:always_on}. To precisely assess their relative effect, we calculate the difference in outcomes produced by changing the QS vs the pipeline. Here's how we obtain such outcome data:
\begin{enumerate}
    \item Take the example of QSes. For each non-random QS $q'$, we list the scores $\delta(f(h, q', d, b, s, n))$ for different values of $h, d, b, s, n$. Since there are four non-random QSes, this gives us four sets of matched observations.
    \item We follow an analogous procedure for prediction pipelines, where we obtain seven matched observation sets. 
\end{enumerate}
A standard method for such analysis is the \emph{Friedman} test \cite{friedman1}, but note here that the number of matched observations for the two cases might be different, which implies different statistical power. Also we might not directly compare the \emph{p-values} since they are a measure of significance.

Instead we use \emph{Kendal's W} to directly measure the \emph{effect size} \cite{Tomczak2014}. These effect sizes for the QS and pipeline parameters respectively are $0.34$ and $0.25$; the effect size here measures agreement, i.e., using different QSes produce similar results (higher agreement), relative to using different pipelines. We also built an \emph{Explainable Boosting Machine (EBM)} \cite{10.1145/2487575.2487579,10.1145/2339530.2339556} on our observations, which is a form of \emph{Generalized Additive Model} that takes into account pairwise interactions. The global feature importance\footnote{The EBM was constructed using different train-test splits, hence both mean and standard deviations across these splits are reported.} for \emph{QS+pipeline}, \emph{QS} and \emph{pipeline} respectively are $0.41 \pm 0.01$, $0.42 \pm 0.02$, $0.63 \pm 0.01$ - which (a) justifies looking at the marginal effects since the importance score for \emph{QS} and \emph{pipeline} independently are at least as large as \emph{QS+pipeline}, and (b) corroborates that changing pipelines has a greater impact.

\subsection{Effect of Batch/Seed Size}
\label{sec:batch_seed_sizes}
We  perform a \emph{Wilcoxon signed-rank test} \cite{wilcoxon} to assess the effect of batch/seed sizes on $\delta$. This is a paired test and ideally we should match observations $\delta(f(h, q, d, 200, 200, n))$ and $\delta(f(h, q, d, 500, 500, n))$. However, recall that since different batch/seed sizes don't lead to the same values of $n$ - we explicitly align the sizes for such comparison (detailed in \S \ref{sec:app_batch_size_avg}).

The overall \emph{p-value} of $0.90$ indicates that our batch/seed settings don't influence $\delta$ in general. The exception is \emph{RoBERTa}, with \emph{p-value}$=1.32e{-10}$. A further one-sided test tells us that the batch/seed size setting of $(200, 200)$ leads to greater $\delta$ values (\emph{p-value}$=6.57e{-11}$).

\begin{table}
  \centering
  \begin{tabular}{lr|lr}

    \hline
  
\emph{Predictor} & \emph{p-value} & \emph{QS} & \emph{p-value} \\
    \hline
    LinSVC-WV& $0.18$ & CAL& $0.77$\\
    LinSVC-USE& $0.41$&  DAL& $0.02$ \\
    LinSVC-MP&$0.60$ & Margin &$0.32$\\
    RF-WV& $0.13$& REAL & $0.07$\\
    RF-USE& $0.03$& \\
    RF-MP& $0.03$& &\\
    RoBERTa& $1.32e{-10}$& &\\
    \hline
    \multicolumn{4}{c}{\textbf{Overall:}  $0.90$}\\
    \hline
  \end{tabular}
  \caption{The \emph{p-values} for a two-sided \emph{Wilcoxon signed-rank test} over $\delta$ values, from using batch/seed size $(200, 200)$ vs $(500, 500)$. See \S \ref{sec:batch_seed_sizes} for details.}
  \label{tab:wilcoxon_batch_seed_sizes}
\end{table}

\subsection{Effect of Representation }
Finally, we assess the effect of text representation on relative improvements. Since we want to evaluate representations alone (the prediction pipeline as a whole was already evaluated in \S \ref{sec:expected_gains}), we ignore \emph{RoBERTa} for this exercise, since its an end-to-end classifier.

\begin{figure}[ht]
\begin{center}
\includegraphics[width=0.45\textwidth]{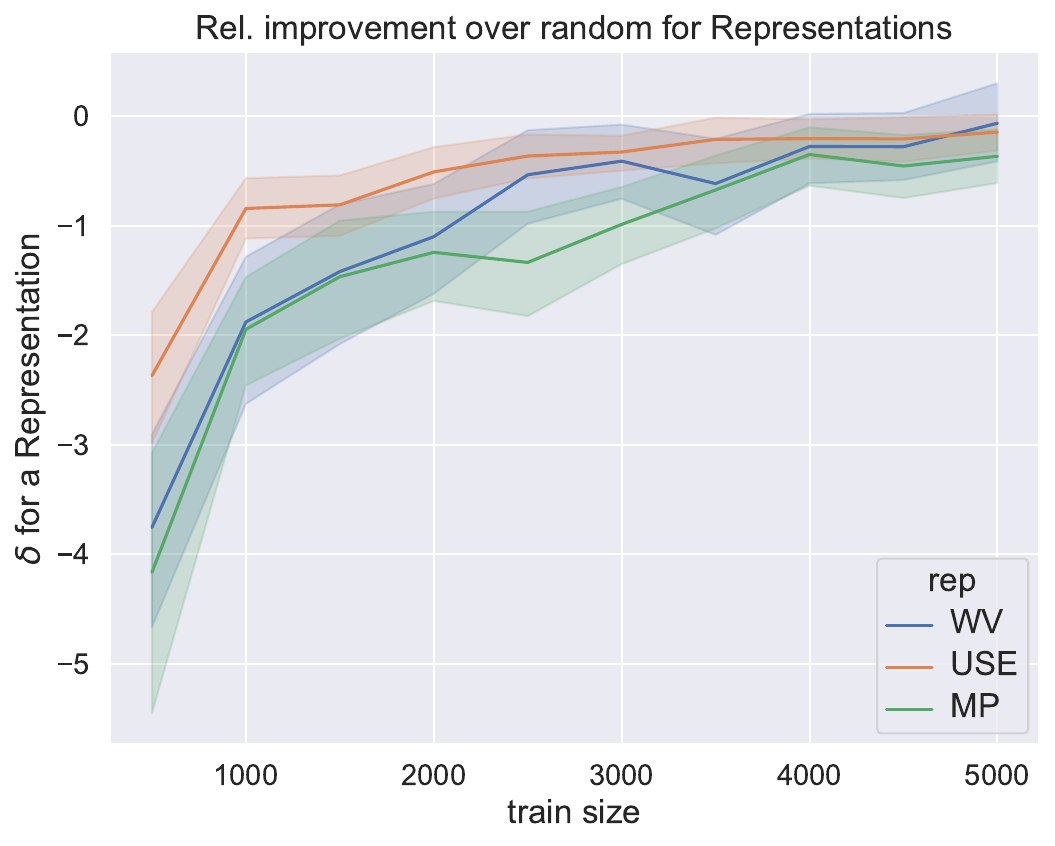}
\caption{Effect of text representations on the relative improvement. }\label{fig:effect_rep}
\end{center}
\end{figure}

Figure \ref{fig:effect_rep} shows how the relative improvement $\delta$
varies with the embedding used, marginalized over other configuration variables.

We note that \emph{USE} outperforms \emph{MPNet}. This is surprising to us because on the MTEB \cite{muennighoff2022mteb} benchmarks \emph{MPNet} scores much higher. A hypothesis that might explain both results is that \emph{USE} doesn't capture fine-grained contexts as much as \emph{MPNet} does; while this might be problematic for MTEB (esp. tasks that rely on precise similarity measurement, such as retrieval), the fuzzier embedding space of \emph{USE} is better in terms of covering the concept space in the dataset earlier in the AL process. This enables better assessment of informativeness, and therefore, sampling, by a non-random QS.

\section{Summary and Conclusion}

After extensive evaluation of different AL algorithms, we are forced to conclude that it is difficult to practically benefit from AL. Gains from QSes are inconsistent across datasets, prediction pipelines and text representations. In fact, between QSes and prediction pipelines, the latter seems to have a greater influence on the relative improvement over random (\S \ref{sec:pipeline_or_QS}). The only general pattern we see is that positive relative improvements become likely as labeled instances accumulate; but these improvements are too small to be broadly useful (\S \ref{sec:expected_gains}). Another reason as to why it is hard to derive any practical advice is that we lack the tools, theoretical or empirical, to identify a settings-specific warm-start size; when do we stop labeling to realize gains, however small?
Further, we noted in \S \ref{sec:always_on} that using AL in an ``Always ON'' mode can actually perform worse than random sampling. 

The use of \emph{RoBERTa} as the prediction pipeline is the only (isolated) case where we see consistent positive relative improvements. Our hypothesis as to why is that an end-to-end classifier has a more coherent view of the overall distribution, and therefore informativeness of a sample. But, obviously, we can't discount the role that \emph{RoBERTa's} specific pre-training might play here, and further experimentation is required to disentangle their respective influences. Even in this case, we point out that (1) it provides further evidence for the argument that the QS  alone does not decide outcomes, and (2) while positive, the gains aren't considerable, with $\delta \approx 1\%$ (see Figure \ref{fig:combined_rel_improve} (a)-(f)).

Although extensive, this study may be considered ``limited'' relative to real-world variances, e.g., many more choices of classifiers, datasets, which leads us to suspect that the true picture is probably more dismal.

What might we do to make the field of AL more useful? 
We feel the biggest problem in AL use is that practitioners have to \emph{blindly guess} what specific AL technique will work best for their problem. 
As a field we need to embrace a broader discourse where the success of a technique needs to be tied to fundamental properties of datasets, e.g., \emph{topological} features \cite{10.3389/frai.2021.667963}, and predictors, e.g., \emph{VC dimension} \cite{Vapnik1995}, that are identifiable in an \emph{unsupervised} manner in novel settings.

\section{Limitations}
Being an empirical work, our conclusions are tied to the algorithms and settings analyzed. In particular, the experiments (a) don't include \emph{Large Language Models}, or (b) is not exhaustive wrt hyperparameters such as batch and seed set sizes; we use two settings, but please note that there is no standard way to select values for these \emph{a priori}. Another aspect that is not considered here is the difference between academic vs real-world datasets, which might lead to different behaviors for a QS \cite{margatina-aletras-2023-limitations}

We also point out that the shortcomings of individual QSes themselves are not a limitation of this study, which maybe seen as an empirical survey with the goal of thoroughly evaluating existing QSes \emph{as-is}.

\section{Acknowledgements}
We thank Sashank Gummuluri for early results in various practical settings. Joshua Selinger reviewed multiple drafts of the paper, and proposed alternative methods of measurement, for which we are grateful to him. We also owe a debt of gratitude to Mandar Mutalikdesai for his encouragement and continued support of this project. The authors would also like to thank reviewers   from the ACL Rolling Review process, especially YZn8, who helped improve various technical aspects of the paper.



\bibliography{refs}

\clearpage
\newpage

\appendix



\section{Pseudo-code for Batch Active Learning}
\label{sec:appendix_pseudocode_bal}

\begin{algorithm}[tbh]
 \KwIn{Unlabeled data $X_{U}$, test data $(X_{test}, Y_{test})$, query strategy $\mathcal{Q}$, seed set selection strategy $\mathcal{A}$, search space $\Theta$ for model $\mathcal{M}$, seed size $s$, batch size $b$, number of iterations $T$, metric $\mathcal{V}$}
 \KwResult{Scores on test set at various iterations $\{(\mathcal{V}_0, 0), (\mathcal{V}_1, 1), ..., (\mathcal{V}_T, T)\}$}
$result \gets \{\}$ \tcp{to be returned}  
$ X_{L,0}, X_{U,0}  \gets \mathcal{A}(X_{U}, s)$ \\
$(X_{L,0}, Y_{L,0})  \gets \text{obtain labels for } X_{L,0} $\\
$M_0 \gets \argmax_{\theta \in \Theta}M_\theta((X_{L,0},Y_{L,0}))$  \tcp{both model selection and calibration are performed}
$\mathcal{V}_0 \gets \mathcal{V}(M_0(X_{test}), Y_{test})$ \\
$result \gets result \cup \{(\mathcal{V}_0, 0 )\}$\\ 
\For {$t\gets1$ \KwTo $T$}{
    $ X_{L,t}^{new}, X_{U,t}   \gets \mathcal{Q}(M_{t-1}, X_{U,t-1}, (X_{L,t-1},Y_{L,t-1}), b)$ \\
    $(X_{L,t}^{new}, Y_{L,t}^{new})  \gets \text{obtain labels for } X_{L,t}^{new} $\\
    $(X_{L,t}, Y_{L,t})  \gets \text{add } (X_{L,t}^{new}, Y_{L,t}^{new}) \text{ to } 
    (X_{L,t-1}, y_{L,t-1}) $\\
    $M_t \gets \argmax_{\theta \in \Theta}M_\theta((X_{L,t},Y_{L,t}))$  
    $\mathcal{V}_t \gets \mathcal{V}(M_t(X_{test}), Y_{test})$ \\
    $result \gets result \cup \{(\mathcal{V}_t, t )\}$\\ 
 }
\Return $result$
 \caption{Batch Active Learning.}
 \label{algo:sketch}
\end{algorithm}

At a high-level, at every AL iteration $1 \leq t \leq T$, we use a query strategy $\mathcal{Q}$ to select a $b$-sized batch of instances from the unlabeled pool of data (line 8). We obtain labels for this set (line 9) and add it to the existing pool of labeled data (line 10). We then train a model $M_t$ over this data (line 11). We emphasize that:
\begin{enumerate}
    \item The model $M_t$ is obtained after performing \emph{model selection} over its hyperparameter space $\Theta$, using \emph{grid-search} against a \emph{validation set}. The validation set is a label-stratified subset (a $20\%$ split) of the current labeled set; the rest is used for training.
    \item The model is also \emph{calibrated}\footnote{A notable exception is in our use of the \emph{RoBERTa} model, which already is well calibrated \cite{desai-durrett-2020-calibration}.}. This is critical since query strategies $\mathcal{Q}$ often use the predicted class probabilities from $M_t$. We use \emph{Platt scaling} \cite{platt_calib,calib_compare}. 
\end{enumerate}

The process is initialized by selecting a seed set of size $s$ from the unlabeled data pool, using a strategy $\mathcal{A}$ (line 2). We use random selection for this step.

We also note that a ``model'' here might mean a combination of a text representation, e.g., \emph{word vectors}, and a classifier, e.g., \emph{Random Forest}; further detailed in Section \ref{sec:method}.

\section{Experiment Configurations}
\label{sec:appendix_experiment_space}

 In our experiments, we vary \emph{classifiers}, \emph{text representations} (we often jointly refer to them as a \emph{prediction pipeline}), \emph{batch size}, \emph{seed size} and, of course, \emph{query strategies}. These combinations are visualized in Figure \ref{fig:experiment_space}, and are detailed in Section.

 These combinations are listed below:
\begin{enumerate}
    \item \textbf{Prediction pipeline}: There are two categories of pipelines we use:
    \begin{enumerate}
        
        \item Separate representation and classifier: The representations used are \emph{USE} \cite{USE}, \emph{MPNet} \cite{mpnet} and \emph{word vectors}\footnote{The vectors of all words in a sentence are averaged to obtain its representation.} (we use the models provided by the \emph{spaCy} library \cite{spacy}). For classification, we use \emph{Random Forests (RF)} \cite{randomforests} and \emph{Support Vector Machines} \cite{svm} with a \emph{linear kernel} - we'll term the latter as ``\emph{LinearSVC}''.  
        
        We use off-the-shelf representations and they are \emph{not} fine-tuned on our data. Only the classifiers are trained on our data.
        
        \item End-to-end classifier: This does not require a separate representation model. We use \emph{RoBERTa} \cite{roberta} (a variant of \emph{BERT}). This is fine-tuned on the labeled data at each AL iteration.
    
    \end{enumerate}

    Hyperparameter search spaces are detailed in Section \ref{sec:app_clf_params} of the Appendix. As noted in Section \ref{sec:bal_overview}, model selection and calibration are performed during training of a prediction pipeline. The only exception is \emph{RoBERTa}, which has been shown to be well-calibrated out of the box \cite{desai-durrett-2020-calibration}.

    The first category gives us $2\times3=6$ combinations. Counting \emph{RoBERTa}, we have $\boldsymbol{7}$ \textbf{prediction pipelines} in our study. 
    
    \item \textbf{Query Strategy}: we list these below, with the year of publication mentioned, to show our focus on contemporary techniques:
    \begin{enumerate}
    \item \emph{Random}: the batch is selected uniformly at random. This forms our baseline.
    \item \emph{Margin}\footnote{Also referred to as \emph{Smallest Margin} or \emph{Breaking Ties}.} \cite{10.1007/3-540-44816-0_31} (2001): this selects instances with the smallest differences between the confidence of the most likely and the second-most likely predicted (by the current classifier\footnote{Note that in reference to Algorithm \ref{algo:sketch}, at iteration $t$, the query strategy $\mathcal{Q}$ uses model $M_{t-1}$.}) classes. Despite being a relatively old technique, it continues to be competitive \cite{schroeder2022-revisiting}. 
    \item \emph{Contrastive Active Learning (CAL)} \cite{cal} (2021): chooses instances whose predicted class-probability distribution is the most different (based on \emph{KL divergence}) from those of their $k$-nearest neighbors. This is similar to another work \cite{XAI_human_in_the_loop}, where such conflicts are detected using the \emph{explanation space} produced by XAI techniques.
        
    \item \emph{Discriminative Active Learning (DAL)} \cite{gissin2019discriminative,ein-dor-etal-2020-active} (2019): a binary classifier (a feedforward neural network) is constructed  to discriminate between labeled and unlabeled data, and then selects unlabeled instances with the greatest predicted probability of being unlabeled. This picks examples that are most different from the labeled instances in this classifier's representation space. While the original work \cite{gissin2019discriminative} only considers image datasets, a separate study shows its efficacy on text \cite{ein-dor-etal-2020-active}.

    \item \emph{Representative Errors for Active Learning (REAL)} \cite{REAL} (2023): identifies clusters in the unlabeled pool and assigns the majority predicted label as a ``pseudo-label'' to all points in it. Instances are then sampled whose predictions differ from the pseudo-label. The extent of disagreement and cluster size are factored into the sampling step
    
    \end{enumerate}
    We use a total of $\boldsymbol{5}$ \textbf{query strategies}.

    \item \textbf{Datasets}: we use $\boldsymbol{5}$ \textbf{standard datasets}: \emph{agnews}, \emph{sst-2}, \emph{imdb}, \emph{pubmed} and \emph{dbpedia-5} (a $5$-label version of the standard \emph{dbpedia} dataset that we created). These are detailed in Table \ref{tab:datasets}. The extent of class imbalance is represented by the \emph{label entropy} column, which is calculated as $\sum_{i\in C} -p_i \log_{|C|} p_i$, with $C$ being the set of classes.
    
    \begin{table*}[ht] 
  \centering
   \resizebox{\textwidth}{!}{\begin{tabular}{p{0.1\linewidth}  p{0.1\linewidth} p{0.1\linewidth}  p{0.65\linewidth}}
    \toprule
    Dataset & \# classes & Label entropy & Description \\
    \midrule
    sst-2 & \centering 2 & 1.0 & Single sentences extracted from movie reviews with their sentiment label \cite{sst2-socher-etal-2013-recursive}.  \\
    imdb & \centering 2  &  1.0 &  Movie reviews with corresponding sentiment label \cite{imbd-maas-EtAl:2011:ACL-HLT2011}. \\
    agnews & \centering 4 & 1.0 &  News articles with their topic category \cite{NIPS2015_250cf8b5}. \\
    pubmed & \centering 4  & 0.9 &  Sentences in medical articles' abstracts which are labeled with their role on the abstract \cite{dernoncourt-lee-2017-pubmed}.\\
    dbpedia-5 & \centering 5  &  1.0 &  A subset of \emph{dbpedia} \cite{NIPS2015_250cf8b5} which contains Wikipedia articles accompanied by a topic label. The original dataset's instances are evenly distributed across $14$ classes. To form \emph{dbpedia-5}, we use only the first $5$ classes:  \emph{Company, EducationalInstitution, Artist, Athlete, OfficeHolder}. This was done to reduce the training time of one-vs-all classifiers, e.g., \emph{LinearSVC}.
   \\
  
    \bottomrule
  \end{tabular}}
    \caption{Datasets used. Label entropy represents class imbalance - see \S \ref{sec:appendix_experiment_space} for description.}  
     \label{tab:datasets}
\end{table*}
    
    \item \textbf{Batch and Seed sizes}: We use batch and seed size combinations of $(200, 200)$ and $(500, 500)$. This is a total of $\boldsymbol{2}$ \textbf{combinations}.

    \item \textbf{Trials}: For statistical significance, we run $\boldsymbol{3}$ \textbf{trials} for each combination of the above settings.
\end{enumerate}


\section{In what data regimes do query strategies most differ?}
\label{sec:app_variance}

\begin{figure}[ht]
\begin{center}
\includegraphics[width=0.45\textwidth]{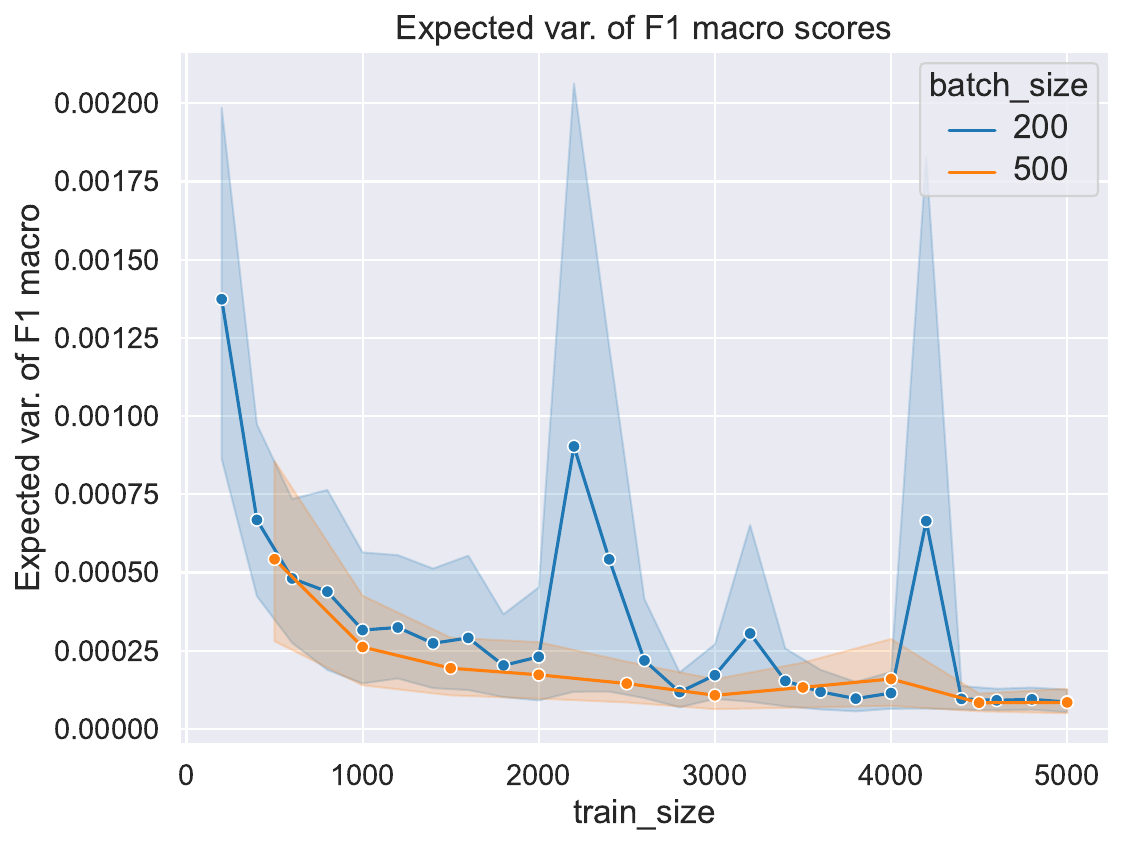}
\caption{Expectation over variance of \emph{F1-macro} given a pipeline and dataset, plotted against size of labeled data. Note that the batch/side sizes don't strongly influence trends.} \label{fig:val_std}
\end{center}
\end{figure}

We would intuitively expect that \emph{F1-macro} scores from different QSes (for a given pipeline and dataset) should converge as we see more data due to at least two reasons: 
\begin{itemize}
    \item  The concept space in the data would be eventually covered after a certain number of instances. Adding more data isn't likely to add more information, i.e., there are \emph{diminishing returns} from adding more data.
    \item  At later iterations, there is less of the unlabeled pool to choose from.
\end{itemize}
Indeed, Figure \ref{fig:val_std} confirms this.  We first compute variances in \emph{F1-macro} scores for each different pipeline/dataset combination\footnote{This step comes first since the accuracies obtained by a \emph{LinearSVC} would be very different from those by \emph{RoBERTa}, and we don't want to mix them.} across QSes at a given labeled set size. And then we average these variances across datasets and pipelines - this is the $y$-axis. We see that the expected variance shrinks after a while, and at $5000$ labeled points it is close to zero, i.e., the differences from using different QSes, pipelines etc isn't much. This is why we restrict the labeled set size to $5000$ instances in our experiments (as mentioned in \S \ref{sec:metrics_other_settings}). 

\section{Reproducibility Experiments}
\label{sec:app_repr}

As mentioned earlier, our intention is \emph{not} to suggest that the techniques we evaluate, e.g., REAL, CAL, DAL, don't work. In the specific settings discussed in their respective papers, they most likely perform as reported. In the interest of fairness, we have conducted limited independent tests that confirm this.

\begin{figure*}[ht]
\begin{center}
    \centerline{\includegraphics[width=0.8\textwidth]{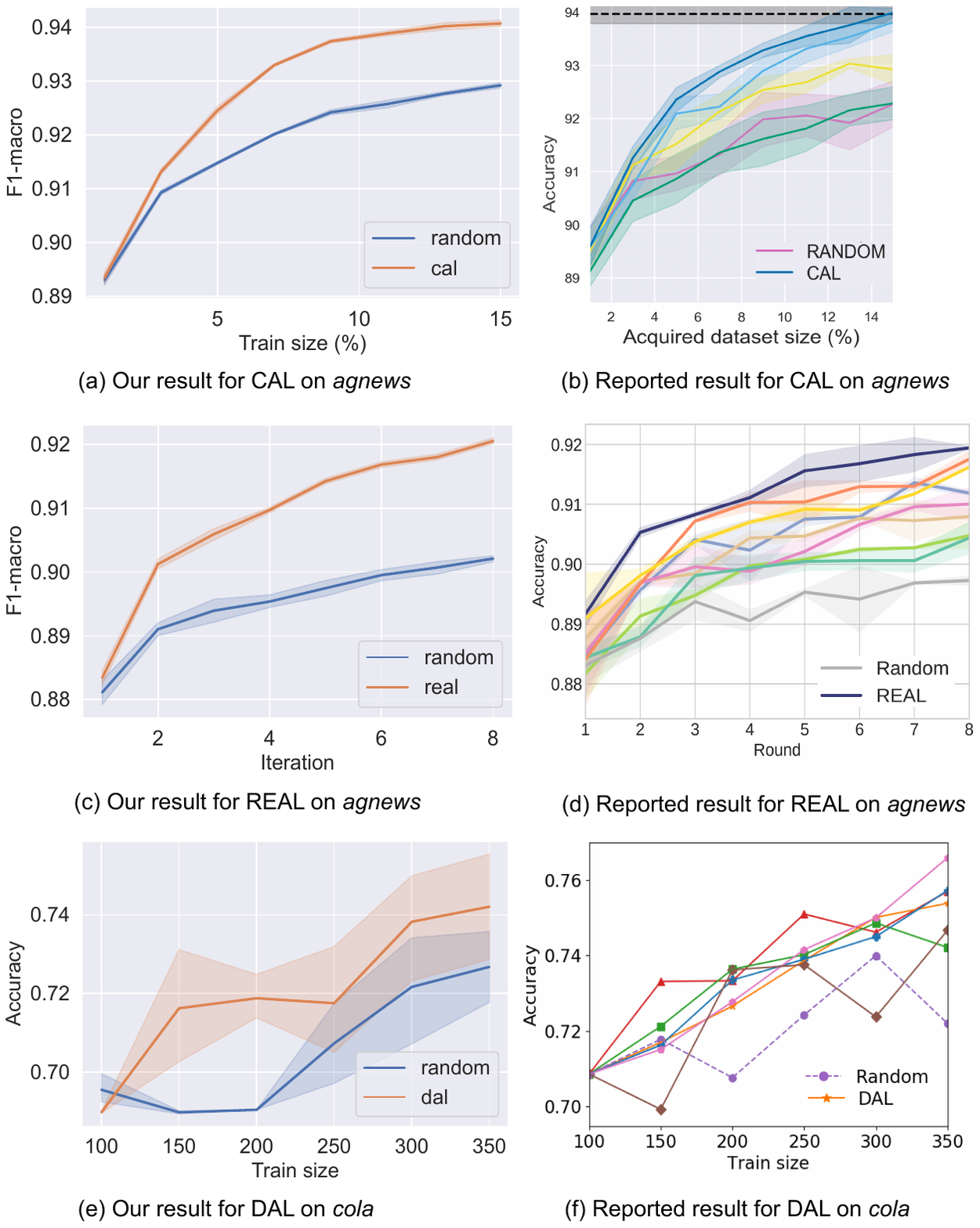}}
    \caption{Comparison between published results in \cite{cal,REAL,ein-dor-etal-2020-active} and ours with the same settings for CAL, REAL, DAL.}
    \label{fig:rep_all}
\end{center}
\end{figure*}

In all cases, we have attempted to replicate the original settings, e.g. same train/development/test data split, model type, seed/batch sizes, number of AL iterations as shown in Table \ref{tab:rep_setting}.  For CAL, REAL, we report the F1-macro scores on \emph{agnews}, in which classes are evenly distributed, instead of the accuracy provided in the original papers. For DAL, we use the dataset \emph{cola}\footnote{https://nyu-mll.github.io/CoLA/} and utilise the \emph{Hugging Face} library to finetune BERT (while the original work employs \emph{TensorFlow}\footnote{https://www.tensorflow.org/}, but we use equivalent settings).
Figure \ref{fig:rep_all} shows a comparison between our results and the reported ones in these papers \cite{cal,REAL,ein-dor-etal-2020-active} for CAL, REAL, DAL, respectively. Despite some minor differences in the setups, we observe that these AL methods work as described in their respective papers in these settings. 

One significant difference between these settings compared to our methodology is the use of a predetermined \emph{labeled} development set for all BERT/RoBERTa model finetuning. This set is relatively larger than the AL batch or seed size and is not part the labeled data available at each AL iteration. This is impractical in scenarios where AL is typically used: labeling is expensive. Moreover, in some cases, there is no model selection performed, which we remedy in our experiments (Section \ref{sec:bal_overview}).

\begin{table*}[t]

  \begin{tabular}{p{0.08\linewidth} p{0.08\linewidth} p{0.1\linewidth} p{0.32\linewidth} p{0.18\linewidth} p{0.1\linewidth}}
  \toprule
    AL & Dataset & AL loop  &  Classifier \& text representation & QS parameters & Metric\\
    \midrule
     CAL & \emph{agnews}  &  b=2280 \newline s=1140   \newline T = 7 &  BERT (bert-base-cased) \newline [CLS] at the last hidden layer 
     \newline learning rate = 2e-5
     \newline train batch size = 16
     \newline \# epochs = 3
     \newline sequence length = 128
     \newline warmup ratio = 0.1
     \newline \# evaluations per epoch = 5
     & \# neighbors=10 & F1-macro\\
    
    \midrule

     DAL & \emph{cola} & b=50 \newline s=100   \newline T = 5  & BERT (bert-base-uncased)  
     \newline [CLS] at the pooled layer 
     \newline learning rate = 5e-5
     \newline train batch size = 50
     \newline \# epochs = 5
     \newline sequence length = 50
     \newline warmup ratio = 0
     \newline \# evaluations per epoch = 1     
     & - & Accuracy\\
    \midrule
     REAL & \emph{agnews} &  b=150\newline s=100\newline T = 8 & RoBERTa (roberta-base) 
     \newline [CLS] at the last hidden layer 
     \newline learning rate = 2e-5
     \newline train batch size = 8
     \newline \# epochs = 4
     \newline sequence length = 96
     \newline warmup ratio = 0.1
     \newline \# evaluations per epoch = 4
     & \# clusters=25 & F1-macro\\    
    \bottomrule
  \end{tabular}
 \caption{Settings for reproducibility experiments.}
     \label{tab:rep_setting}
\end{table*}

\section{Hyperparameters}
\label{sec:app_hyp}
\subsection{Query Strategy (QS) hyperparameters}
\label{sec:app_qs_params}
For each QS's hypeparameters, we use the values recommended by the authors in corresponding papers. This means setting number of nearest neighbors in CAL to 10, number of clusters in DAL to 25, and keeping the same discriminative model in REAL.

\subsection{Hyperparameters search for prediction pipelines}
\label{sec:app_clf_params}
Table \ref{tab:clf_params} shows the search space for hyperparameters we use for each classifier.
\begin{table*}[ht]

\centering
  \begin{tabular}{p{0.3\linewidth} p{0.5\linewidth} }
  \toprule
   Classifier & Hyperparameters \\
   \midrule
    RoBERTa & roberta-base 
     \newline [CLS] at the last hidden layer 
     \newline learning rate = \{3e-5, 5e-5\} 
     \newline train batch size = 16
     \newline \# epochs = \{5, 10\}
     \newline sequence length = 128
     \newline warmup ratio = 0.1
     \newline \# evaluations per epoch = 5
     \\  
     \midrule
    LinearSVC & C = \{0.001, 0.01, 0.1, 1, 10, 100, 1000\}
    \newline class weight = balanced 
    \\
    \midrule
    RF &  min samples leaf = \{1, 5, 9\}
    \newline  \# estimators = \{5, 10, 20, 30, 40, 50\}
    \newline max depth = \{5, 10, 15, 20, 25, 30\}
    \newline class weight = balanced 
    \newline max features = sqrt
    \\
 \bottomrule
  \end{tabular}
\caption{Hyperparameters for each classifier in the prediction pipelines.} 
\label{tab:clf_params}
\end{table*}

\section{Averaging over Different Batch-Sizes}
\label{sec:app_batch_size_avg}
When computing expectations over different batch/seed sizes (like in Equation \ref{eqn:avg_gain}) a challenge is that different settings don't lead to same number of instances. For ex., for  $b=200, s=200$, the size of the trained pool assumes the values $200, 400, .. , 5000$, and for $b=500, s=500$,  the sizes are $500, 1000, .., 5000$. To compute an expectation of the form $\mathbb{E}_{b,s}[., n']$, we use the sizes from the larger batch, i.e.,  $n' \in \{500, 1000, .., 5000\}$, and map the \emph{closest sizes} from the smaller batch to them. For ex., here are some size mappings from the small batch case to the larger one: $800 \rightarrow 1000, 1000\rightarrow1000, 1200\rightarrow 1000, 1400\rightarrow 1500, 1600 \rightarrow 1500$.

\section{Always ON Mode}
\label{sec:app_always_on}
Table \ref{tab:always_on_std} presents standard deviations for the ``Always ON'' case, and is a companion to Table \ref{tab:always_on} in \S \ref{sec:always_on}. Note the extremely high variances in moving across combinations of the configurations and size of the labeled set.

\begin{table*}[!ht]
    \centering
    \begin{tabular}{lrrr}
    \toprule
        \emph{Avg. for} & \% times $\delta < 0 $ & $\overline{\delta}_{\geq 0}$ & $\overline{\delta}$ \\ \toprule
        \textbf{Overall} & 51.82 & 0.89 $\pm$ 0.92& -0.74 $\pm$ 3.02\\ \midrule
        LinSVC-WV & 61.71 & 0.70  $\pm$ 0.60 & -1.90  $\pm$ 3.94\\ 
        LinSVC-USE & 61.57 & 0.46 $\pm$ 0.49 & -0.64  $\pm$ 1.85\\ 
        LinSVC-MP & 63.71 & 0.40 $\pm$ 0.44 & -1.48 $\pm$ 3.53 \\ 
        
        RF-WV & 47.29 & 1.31 $\pm$ 1.01 & -0.30 $\pm$ 2.63 \\ 
        RF-USE & 60.57 & 0.71  $\pm$ 0.69 & -0.63 $\pm$ 1.85 \\ 
        RF-MP & 60.14 & 0.60 $\pm$ 0.55 & -1.24 $\pm$ 3.59 \\ 
        
        RoBERTa & 7.71 & 1.29 $\pm$ 1.17 & 1.01 $\pm$ 1.94 \\ 
        \midrule
        cal & 55.60 & 0.81 $\pm$ 0.86 & -1.07 $\pm$ 3.23 \\ 
        dal & 70.12 & 0.82 $\pm$ 0.94 & -1.29 $\pm$ 3.22 \\ 
        margin & 38.45 & 0.97 $\pm$ 0.88 & -0.25  $\pm$ 2.78 \\ 
        real & 43.10 & 0.89 $\pm$ 0.99 & -0.34  $\pm$ 2.67\\ \bottomrule
    \end{tabular}
    \caption{The $\%$-age of times model \emph{F1-macro} scores are worse than random, the average $\delta$s when scores are at least as good as random and average $\delta$s in general. These are identical to the values in Table \ref{tab:always_on} in \S \ref{sec:always_on}, but the standard deviations are additionally shown here.}
\label{tab:always_on_std}
\end{table*}

\end{document}